\documentclass[letterpaper, 10 pt, conference]{ieeeconf}  

\IEEEoverridecommandlockouts                              

\usepackage{graphics} 
\usepackage{epsfig} 
\usepackage{mathptmx} 
\usepackage{times} 
\usepackage{amsmath} 
\usepackage{amssymb}  
\usepackage{algorithm}
\usepackage{algorithmic}
\usepackage{pifont}

\title{\LARGE \bf
RSC: Decentralized Rigid Formation Flocking for Large-Scale Swarms via Hybrid Predictive Control and Online Reconfiguration
}

\author{Ganyu Zou, Linhan Wang, Chen Dai, Siji Chen and Chang-Tien Lu}

\begin{document}

\maketitle
\thispagestyle{empty}
\pagestyle{empty}

\begin{abstract}
Decentralized rigid formation flocking requires a swarm of autonomous agents to maintain a predetermined geometric configuration while moving, relying solely on local sensing and communication. However, existing decentralized control methods struggle to maintain strict inter-agent distance constraints in cluttered environments, often suffering from local minima deadlocks, high frequency control oscillations, or limited flexibility during obstacle navigation, resulting in low success rate. To address these limitations, we propose Rigid Swarm Control (RSC), a decentralized control framework for large-scale rigid formation flocking. To escape local minima via robust long-term planning while ensuring short-term safety, RSC integrates finite-horizon trajectory predictions with a reactive artificial potential field (APF) safety controller within a hybrid architecture. Furthermore, to accelerate formation reassembly after obstacle traversal without interrupting task execution, RSC introduces an online leader-follower reconfiguration mechanism based on stable role exchange. Extensive evaluations in challenging cluttered environments with \(25\) UAVs demonstrate that RSC reliably unifies rigid formation maintenance, obstacle avoidance, and target tracking. Under strict success criteria—collision-free operation with a maximum relative edge-length error below \(10\%\), RSC achieves an \(83\%\) success rate, significantly outperforming existing heuristic and learning-based baselines that fall below \(5\%\).
\end{abstract}

\section{INTRODUCTION}
Flocking refers to the coordinated motion of animals or autonomous agents (e.g., birds, fish, and robots), where individuals follow local interaction rules to achieve a shared group objective. Although classical flocking formulations primarily emphasize cohesion and alignment, many large-scale animal groups in nature also maintain recognizable geometric patterns, such as the V-shaped formations of migrating birds or circular milling patterns in schools of fish. 
This observation suggests that effective coordination inherently involves both \textbf{motion} and \textbf{shape}. 
In robotic applications, maintaining spatial layouts during motion is critical for complex multi-agent missions, particularly in large-scale swarm deployments. Consistent formations enable predictable sensing footprints for cooperative search and area coverage, maintain fixed sensor baselines for multi-view perception and localization, and support coordinated maneuvers in escorting or perimeter patrol tasks.
Motivated by these insights, we study rigid formation flocking, in which a swarm maintains a predetermined geometric configuration by enforcing a set of inter-UAV distance constraints, thereby preserving global shape during motion. 
We consider a decentralized setting where each UAV computes its control using only onboard sensing and communication with first-order neighbors, without access to global information or requiring a central coordinator.

\begin{figure}[t]
  \centering
  \includegraphics[width=\columnwidth]{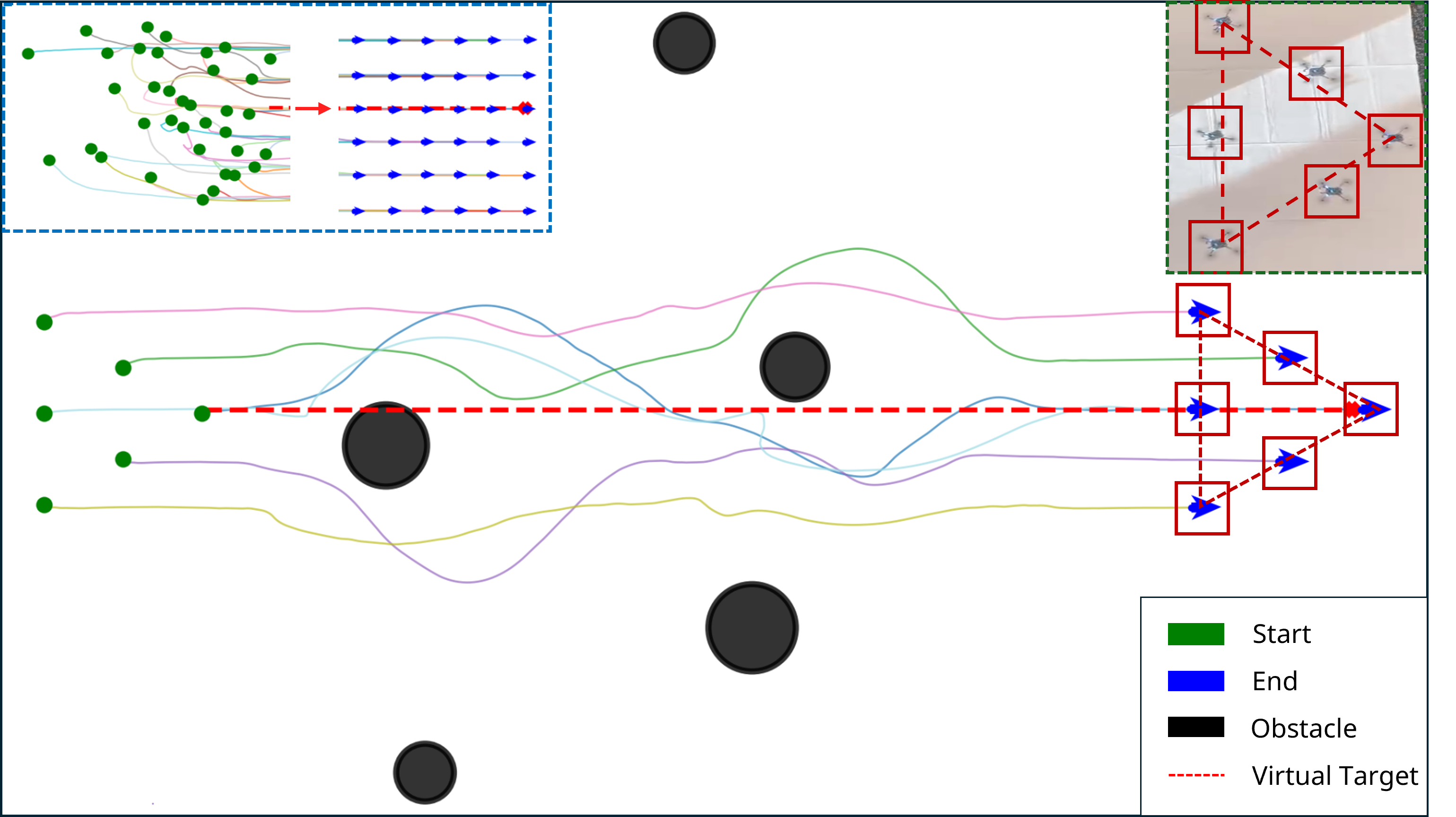}
  \caption{Obstacle avoidance and rigid formation control of six UAVs via the RSC framework. the top-right inset shows the real-world UAV formation. The top-left inset demonstrates RSC controlling a larger UAV array.}
  \label{fig:small_teaser}
  \vspace{-10pt}
\end{figure}
Flocking control for multi-agent systems encompasses heuristic~\cite{olfati-saber_flocking_2006,olfati-saber_consensus_2004,pan_improved_2022}, learning-based~\cite{tolstaya_learning_2017,lee_learning_2022}, and optimization-based approaches. A critical bottleneck in decentralized multi-agent navigation is the susceptibility to \textbf{local minima} in cluttered environments. To mitigate this, heuristic methods often rely on auxiliary mechanisms, such as wall-following optimizations~\cite{kim_escaping_2025}. Meanwhile, learning-based methods aim to bypass local minima using enhanced policies derived from imitation learning~\cite{moukhlis_towards_2025}, reinforcement learning exploration~\cite{zhang_reinforced_2023}, or advanced architectures like Dynamic Graph Neural Networks (DGNNs)~\cite{li_test-time_2025} and Spatial-Temporal GNNs~\cite{chen_learning_2024} that leverage rich topological information. Despite these advances in local decision-making, purely reactive strategies remain fundamentally insufficient for complex environments. Optimization-based methods, such as Model Predictive Control (MPC)~\cite{kuriki_formation_2015,zhu_heuristic_2025,torrente_data-driven_2021}, address this limitation by offering explicit predictive capabilities to avoid local minima; however, their severe computational overhead renders them intractable for large-scale flocking. Consequently, we observe that the fundamental solution lies in endowing agents with \textbf{global predictive and planning capabilities} in a scalable manner.

To maintain precise spatial arrangements, various formation control strategies have been developed, including the leader-follower framework\cite{meng_ji_leader-based_2006,de_souza_neto_decentralized_2019,zhao_uav_2017}, the virtual structures framework\cite{askari_uav_2015}, the Voronoi partition algorithm\cite{hu_formation_2020} and the Markov Decision Process (MDP)-based formulations \cite{azam_uav_2021}. Among these, the leader--follower paradigm is widely adopted in large-scale swarm operations due to the inherent stability and structural simplicity of the approach. To tackle complex missions that involve formation keeping, obstacle avoidance, and target tracking, this paradigm is increasingly integrated into modern learning-based frameworks \cite{jiang_end--end_2023,qiu_drl_2021}. Despite these advancements, research on the rigid formation control of large-scale swarms (e.g., $N > 20$) remains limited, as the performance of most existing methods degrades sharply as the swarm size increases. The primary difficulties at such scales are twofold: first, \textbf{satisfying rigid inter-agent distance} constraints with high accuracy; and second, addressing \textbf{the lack of flexibility at scale}, where local disturbances and complex environmental constraints necessitate rapid and distributed topological reconfiguration.

To address these limitations, we propose \textbf{Rigid Swarm Control (RSC)}, a decentralized control framework for large-scale rigid formation flocking in cluttered environments. RSC integrates finite-horizon trajectory predictions with APF reactive safety control and online formation reconfiguration, enabling each UAV to maintain rigid distance constraints using only local information.
Our framework consists of three components. \textbf{(1)}\textbf{finite-horizon trajectory predictions}: RSC outputs finite-horizon trajectory predictions rather than directly regressing per-step control commands. Per-step controls are then derived under explicit feasibility constraints (e.g., velocity and acceleration bounds). This design mitigates short-sightedness and high-frequency oscillations that are common in one-step regression and improves the learnability of long-horizon coordination. \textbf{(2)}\textbf{Hybrid control}: RSC fuses look-ahead guidance from trajectory prediction with the fast near-field reactivity of an APF-style safety controller. APF provides immediate collision avoidance, while predicted trajectories serve as a goal- and formation-aware guidance signal that helps steer away from APF local minima and potential deadlocks. \textbf{(3)}\textbf{Online leader-follower reconfiguration}: RSC introduces a role-exchange mechanism that updates leader--follower relations online, enabling rapid reconfiguration and formation reassembly after obstacle traversal without interrupting task execution.

We evaluate the method in a challenging setting (\(N=25\) with \(3\)--\(6\) randomly distributed obstacles and random initial UAV positions). We define success as collision-free operation while maintaining a maximum relative edge-length error below \(10\%\). Under this criterion, existing learning-based decentralized baselines and APF-based methods achieve success rates below \(5\%\), whereas RSC attains an \(83\%\) success rate. These results indicate that RSC can reliably accomplish the composite task of decentralized rigid formation, obstacle avoidance, and target tracking.

Our contributions can be summarized as follows:
\begin{itemize}
  \item \textbf{Scalable Decentralized Control Framework (RSC):} We propose a control framework designed for large-scale multi-agent systems that unifies rigid formation maintenance, obstacle avoidance, and target tracking under local communication constraints.
  \item \textbf{Finite-Horizon Trajectory Prediction:} We incorporate predictive trajectory planning, improving control smoothness and the learnability of complex swarm behaviors.
  \item \textbf{Hybrid Control Mechanism:} We fuse finite-horizon predictive guidance with a reactive APF safety controller to effectively mitigate local minima and prevent deadlocks in cluttered environments.
  \item \textbf{Online Leader-Follower Reconfiguration:} We introduce a multi-hop message-passing and role-exchange mechanism to dynamically reconfigure leader-follower relations, accelerating stable formation reassembly during obstacle traversal.
\end{itemize}

\section{METHODOLOGY}
This work focuses on decentralized control for large UAV swarms under local communication, targeting tracking, obstacle avoidance, and rigid formation reassembly. To achieve this, we propose an imitation-learning-based GNN finite-horizon trajectory prediction framework. This section is organized as follows:

In \textbf{Section~\ref{sec:export_controller}}, we construct a centralized APF expert to generate high-quality demonstration data, and define a decentralized APF safety controller for near-field reactivity.

In \textbf{Section~\ref{sec:Learning Policy}}, we detail the decentralized learning policy. We introduce a multi-hop GNN that leverages both the communication graph and the leader--follower tree. Instead of regressing per-step commands, the model predicts finite-horizon trajectories, which are then dynamically clipped and fused with the safety APF to ensure feasibility and smoothness.

In \textbf{Section~\ref{sec:Online Leader--Follower Reconfiguration}}, we present the online leader--follower reconfiguration strategy. We design a lightweight swap operator that dynamically rewires the tree topology based on local information, preventing agent lagging and ensuring rapid formation reassembly in cluttered environments.

\subsection{Expert Controller and Safety Controller}
\label{sec:export_controller}
\paragraph{Expert Algorithm }
Inspired by Chen et al\cite{chen_learning_2024} and Zhao et al\cite{zhao_uav_2017}, We construct a centralized expert policy: at each time step, it uses the global state to output an acceleration command \(a_i\) for UAV \(i\), and linearly combines three control components—formation keeping, obstacle avoidance, and target (virtual-leader) following:
\begin{equation}
a_i = w_f a_i^{\mathrm{form}} + w_o a_i^{\mathrm{obs}} + w_g a_i^{\mathrm{goal}} .
\end{equation}
Additionally, the expert applies the Online Leader--Follower Reconfiguration algorithm described in Section~\ref{sec:Online Leader--Follower Reconfiguration}.
Here, \(a_i^{\mathrm{form}}\) is derived from an improved artificial potential field (APF) for formation control: a structurally constrained attractive term drives UAVs to converge to the desired relative configuration, while a Morse-type repulsive potential and a damping term are incorporated to maintain a minimum safety separation and suppress oscillations. The obstacle-avoidance term \(a_i^{\mathrm{obs}}\) adopts the \(\beta\)-robot projection idea, which models each obstacle as a virtual agent that exerts a repulsive effect to achieve collision-free navigation. The goal-following term \(a_i^{\mathrm{goal}}\) is a PD-like tracking component with respect to the virtual leader/target to accomplish trajectory following. The final command is saturated under velocity/acceleration constraints and then sent to the low-level controller for execution.
\paragraph{Safety Controller.}
We implement a decentralized APF-based safety controller that makes decisions using only the local observations defined in Section~\ref{sec:Local Observation and Aggregation} (i.e., without access to global state). Its control structure follows the same APF formulation as the expert policy, but all formation-maintenance potentials are removed; the controller therefore retains only the reactive obstacle-avoidance and goal/virtual-leader tracking terms to produce a safety-oriented command  \(\mathbf a_{i,\mathrm{APF}}^{h}\). As described in Section~\ref{sec:Finite Horizon Trajectory Prediction with APF Fusion}.

\subsection{Learning Policy}
\label{sec:Learning Policy}

\begin{figure*}[t]
  \centering
  \includegraphics[width=1.00\textwidth]{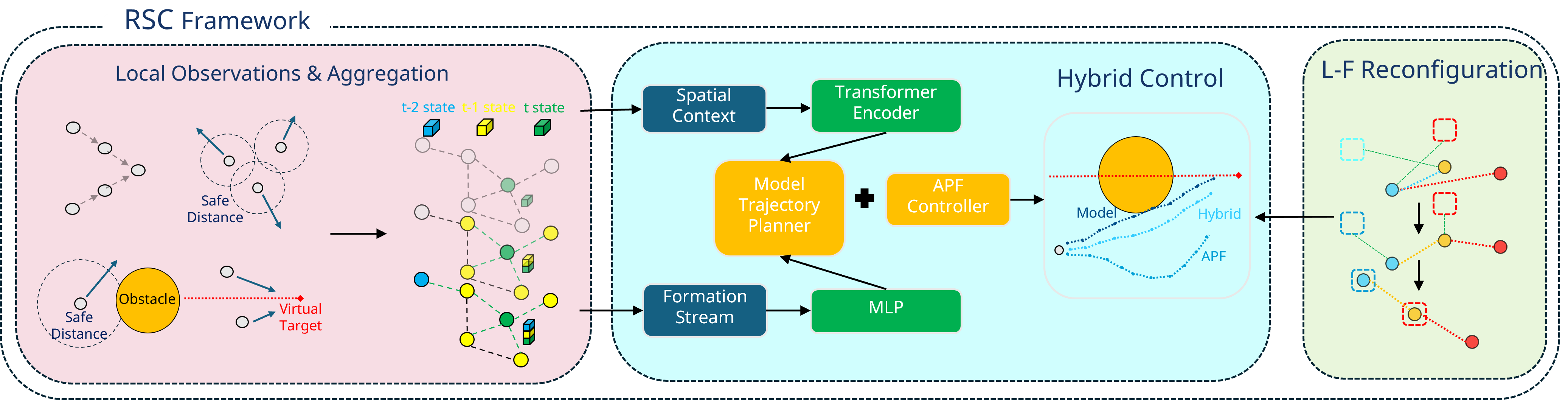}
  \caption{Overview of the RSC framework. Multi-hop aggregation of local observations is utilized as the model input. A hybrid architecture consisting of a transformer encoder\cite{vaswani_attention_2023} and an MLP enables finite-horizon trajectory prediction. The predicted trajectory is combined with an APF safety controller to generate hybrid control. Additionally, an online leader-follower reconfiguration mechanism is introduced to enhance formation flexibility.}
  \label{fig:structure}
\end{figure*}

In this section, we sequentially introduce the design of local observations, the information flow over the communication graph, the imitation-learning mechanism for finite-horizon trajectory prediction with APF fusion, and the online leader--follower reconfiguration. The processes shown in Figure~\ref{fig:structure}.
\subsubsection{Local Observation and Aggregation}
\label{sec:Local Observation and Aggregation}
The local observation of each UAV is restricted to \emph{local relative} information, including the relative displacement and relative velocity with respect to neighboring robots, obstacles, and a virtual leader. The local observation of obstacles is defined by \(r_{i}^k<R_c\), where \(R_c\) is the sensing radius and \(r_{i}^k\) is the distance between robot \(i\) and the projection of robot \(i\) on the \(k\)-th obstacle.

The leader--follower structure is pre-specified: each follower has a unique leader, thus forming a tree topology\cite{tanner_leader--formation_2004,desai_modeling_2001}. At time step \(t\), the \(i\)-th UAV broadcasts all its observation information at time \(t\), denoted by \(X_{i}^{(t)}\), to all neighbors within its communication range. Meanwhile, it receives the observation information broadcast by all neighbors $\mathrm{N}_{i}$ at time \(t-1\) and processes it in two ways.
(1) For all received messages \(X_{j}^{(t-1)}\), we perform mean aggregation:
\begin{equation}
    {X'}_{i}^{(t)}=\frac{1}{\left|\mathrm{N}_{i}^{(t)}\right|}
    \sum_{j\in \mathrm{N}_{i}^{(t)}} X_{j}^{(t-1)} .
\end{equation}
(2) For formation information, we additionally store the relative displacements between UAV \(i\) and all its neighbors, \(\mathbf{p}_{i,j}\) for \(j\in \mathrm{N}_{i}^{(t)}\), denoted by \({P'}_{i}^{(t)}\).

Assuming each UAV can memorize historical information with a length of \(L\), we apply the same aggregation scheme as in (1) for \(L\) rounds. The final mean-aggregated information is
\begin{equation}
    X_{i_s}^{(t)}=\Big[\,X_{i}^{(t)}\;\Big\|\_{l=1,\ldots,L}\;{X'}_{i}^{(l)}\,\Big].
\end{equation}
For formation information, we construct the communication  $G_c$ using all nodes and edges, and use the following equation to enable follower \(i\) to obtain the relative position of its leader \(\ell(i)\) within at most \(L\) hops:
\begin{equation}
{\mathbf p}_{i\to \ell(i)}(t)=
\sum_{m=0}^{d-1} \mathbf p_{v_m,v_{m+1}}(t-m).
\end{equation}
where \(d\) is the length of the shortest communication path between UAV \(i\) and its leader \(\ell(i)\) in the communication graph. we compute the target positions induced by the current formation offsets $\mathbf p_{\ell(i)\to tar(i)}$ (for roots, the target is the virtual target). 
\begin{equation}
    {\mathbf p}_{i\to tar(i)}(t)=\mathbf p_{\ell(i)\to tar(i)}(t)+{\mathbf p}_{i\to \ell(i)}(t).
\end{equation}
\subsubsection{Finite Horizon Trajectory Prediction with APF Fusion}
\label{sec:Finite Horizon Trajectory Prediction with APF Fusion}
Most decentralized flocking algorithms regress per-step control commands. However, our experiments reveal that such myopic approaches lack the long-term planning necessary for complex coordination in cluttered environments. Inspired by motion planning in autonomous driving\cite{bansal_chauffeurnet_2019,hu_planning-oriented_2023,zeng_end--end_2021}, we shift from single-step reactive control to finite-horizon trajectory prediction. Consequently, we design the model output as a local trajectory sequence of length \(H\) time steps, \(\{\mathbf r^{h}_{i}\}_{h=0}^{H-1}\), where \(\mathbf r^{h}\) denotes the relative displacement at step \(h\) expressed in the local frame whose origin is the current UAV position. Accordingly, the absolute target position at step \(h\) is
\begin{equation}
\mathbf p^{h}=\mathbf p^{0}+\mathbf r^{h},
\qquad \mathbf p^{0}=\mathbf p(t),\ \mathbf v^{0}=\mathbf v(t).
\end{equation}
Since the time increment \(\Delta t\) between consecutive steps is small, we approximate each step by constant-acceleration
motion and discretize the dynamics using the trapezoidal relation
\(\mathbf p_{\mathrm{req}}^{h}\approx \mathbf p^{h}+\tfrac{\Delta t}{2}(\mathbf v^{h}+\mathbf v^{h+1})\).
By inverse dynamics that enforces reaching \(\mathbf p_{\mathrm{req}}^{h}\) within one step, the (unconstrained) desired
acceleration at step \(h\) is
\begin{equation}
\mathbf a_{\mathrm{req}}^{h}
=\frac{2}{\Delta t^{2}}\Big(\mathbf p_{\mathrm{req}}^{h}-\mathbf p^{h}-\mathbf v^{h}\Delta t\Big),
\qquad h=0,1,\ldots,H-1.
\end{equation}

\paragraph{Hybrid control (APF fusion).}
To improve safety in near-collision situations, we introduce a distance-based blending weight \(w_i^{h}\in[0,1]\) and
form a convex combination of the tracking-induced acceleration \(\mathbf a_{i,\mathrm{req}}^{h}\) and the decentralized
APF safety controller output \(\mathbf a_{i,\mathrm{APF}}^{h}\).
For UAV \(i\), we define the nearest agent distance as $r_{i}^{\mathrm{agent},h}$, the nearest obstacle distance as $r_{i}^{\mathrm{obs},h}$,
the corresponding risk weight and obstacle-induced risk are
\begin{equation}
w_{i}^{\mathrm{agent},h}
=max \!\left(1-\frac{r_{i}^{\mathrm{agent},h}}{r_{\mathrm{th}}},0\right),
\label{eq:risk_agent_w}
\end{equation}
\begin{equation}
w_{i}^{\mathrm{obs},h}
=max \!\left(1-\frac{r_{i}^{\mathrm{obs},h}}{r_{\mathrm{th}}},0\right),
\label{eq:risk_agent_w}
\end{equation}
and we set
\begin{equation}
w_i^{h}=\max\!\left(w_{i}^{\mathrm{agent},h},\,w_{i}^{\mathrm{obs},h}\right),
\label{eq:risk_final_w}
\end{equation}
The hybrid acceleration command is then given by
\begin{equation}
\mathbf a_{i,\mathrm{hyb}}^{h}
=(1-w_i^{h})\,\mathbf a_{i,\mathrm{req}}^{h}+w_i^{h}\,\mathbf a_{i,\mathrm{APF}}^{h}.
\label{eq:hyb_acc}
\end{equation}

\paragraph{Unified dynamics constraints and rollout.}
We then apply unified dynamic constraints to the hybrid command and roll out the states recursively. We define the
ball-clipping projection operator
\begin{equation}
\Pi_{\rho}(\mathbf{x})
:=
\mathbf{x}\cdot \min\!\left(1,\frac{\rho}{\|\mathbf{x}\|_2}\right).
\label{eq:proj}
\end{equation}
For each prediction step \(h\) and each UAV \(i\), we compute
\begin{align}
\mathbf a_{i}^{h}
&=\Pi_{a_{\max}}\!\left(\mathbf a_{i,\mathrm{hyb}}^{h}\right),
\\
\mathbf v_{i}^{h+1}
&=\Pi_{v_{\max}}\!\left(\mathbf v_{i}^{h}+\mathbf a_{i}^{h}\Delta t\right),
\label{eq:dyn_av}\\
\mathbf p_{i}^{h+1}
&=\mathbf p_{i}^{h}+\frac{\Delta t}{2}\left(\mathbf v_{i}^{h}+\mathbf v_{i}^{h+1}\right).
\label{eq:dyn_p}
\end{align}
When \(w_i^{h}=0\), the controller reduces to pure trajectory tracking; when \(w_i^{h}\to 1\), it smoothly transitions
to the APF safety controller.
\subsection{Online Leader--Follower Reconfiguration}
\label{sec:Online Leader--Follower Reconfiguration}

In preliminary experiments with large swarms, we observed a critical failure mode under fixed topologies: designated leaders frequently fall behind their followers when navigating through cluttered regions. Upon exiting the obstacle zone, the rest of the formation often completes its assembly. Consequently, the lagging leader becomes physically blocked, unable to overtake its followers due to strict inter-agent collision avoidance constraints, which ultimately causes formation failure. 

To address this deadlock and improve formation convergence, we introduce a lightweight \emph{swap} operator that locally rewires the tree by exchanging a leader and its follower. In a decentralized setting, the leader makes swap decisions based on locally available (possibly delayed) information. If a swap is accepted, the leader broadcasts the swap command; all agents that receive it apply the same tree update.

\paragraph{Trigger (distance-based).}
For agent \(i\), a swap \(i \leftrightarrow \ell(i)\) is triggered if both agents would be closer (by a margin \(\delta\)) to the \emph{other} agent's target, i.e.,
\[
\begin{aligned}
\|\mathbf{p}_{i\to \mathrm{tar}[\ell(i)]}\| + \delta &< \|\mathbf{p}_{\ell(i)\to \mathrm{tar}[\ell(i)]}\|, \\
\|\mathbf{p}_{\ell(i)\to \mathrm{tar}(i)}\| + \delta &< \|\mathbf{p}_{i\to \mathrm{tar}(i)}\|.
\end{aligned}
\]
We apply a greedy strategy and perform one call at a time.

\paragraph{Feasibility (connectivity constraint).}
Since swapping changes multiple parent--child relations, we require all affected agents to remain connected in the current communication graph \(G_c\) defined in the previous section. The affected set includes the swapped pair, the old parent of \(\ell(i)\) (if any), the other children of \(\ell(i)\), and all children of \(i\). The leader checks whether the induced subgraph \(G_c[A]\) is connected; only then the swap is allowed.

\paragraph{Broadcast, reconnection, and cooldown.}
If allowed, the leader broadcasts the swap so that all receiving agents modify the tree consistently. To enable agents that were temporarily outside the communication graph to update immediately after reconnection, each committed swap is associated with a monotonically increasing tree version; a reconnected agent obtains the latest version (and the corresponding updates or the current tree) from any neighbor and then updates its local tree. After a tree update, the initiating leader enters a cooldown period and does not issue swap commands to prevent chattering.

\paragraph{Topology update.}
Let \(u\) be the old parent of \(\ell(i)\). After swapping: (i) \(i\) takes over \(u\) (if \(u\neq\emptyset\)); (ii) all children of \(\ell(i)\) except \(i\) are reattached to \(i\); (iii) \(\ell(i)\) becomes a child of \(i\); (iv) the original subtree of \(i\) is transferred to \(\ell(i)\). This preserves a valid tree while changing hierarchy locally.

\section{SIMULATION AND EXPERIMENT}
In this section, we describe the configuration of our RSC model, the baseline methods, the evaluation metrics, and the results from both simulation and real-robot experiments.
\subsection{RSC Model Settings}
\label{sec:RSC Model Settings}
The RSC policy takes as input the agent feature vector \(X_i \in \mathbb{R}^{18}\) and the relative position to its assigned leader \(P_{i\rightarrow \ell(i)} \in \mathbb{R}^{2}\), as defined in Sec.~II-B1. We adopt a two-stream architecture to decouple global coordination from local formation keeping.

\paragraph{Spatial-context stream.}
\(X_i\) is first encoded by a two-layer MLP with hidden dimension 256. We then apply a single GraphSAGE convolution layer (hidden dimension 256) to aggregate neighborhood information according to Eq.~(4). To model temporal context, we stack the resulting node embeddings over the past \(L=3\) time steps into a sequence, which is processed by a Transformer encoder (one layer, \(n_h=4\) heads, feed-forward dimension \(d_{ff}=256\)) to extract spatial flocking-related context.

\paragraph{Local-formation stream.}
\(P_{i\rightarrow \ell(i)}\) is encoded by a two-layer MLP with hidden dimension 256, using LeakyReLU activations and dropout regularization. This stream does not perform neighborhood aggregation to preserve stable relative-geometry features for formation keeping.

\paragraph{Trajectory head and execution.}
The outputs of the two streams are concatenated into a 512-dimensional latent vector and fed into a two-layer MLP head to predict the finite-horizon trajectory in Sec.~II-B2, denoted by \(\textbf{r}_i \in \mathbb{R}^{H\times 2}\) . The predicted trajectory is integrated with the hybrid controller (APF fusion) and rolled out for \(H\) steps under the unified dynamic constraints.

\paragraph{Training with DAgger.}
We train the policy inspired by DAgger~\cite{ross_reduction_2011}. After each policy query, the simulator executes either the expert trajectory or the policy trajectory with probabilities \(\alpha\) and \(1-\alpha\), respectively, where
\[
  k = \max(n_{\mathrm{ep}} - 1000,\; 0), \quad \alpha = \max(0.99^{\,k},\; 0.1),
\]
$n_{\mathrm{ep}}$ denotes the current training epoch number.
We store all collected samples in a replay buffer (aggregated dataset) and update the network using mini-batch training by minimizing the \(\ell_2\) loss between the predicted and expert trajectories with the Adam optimizer~\cite{kingma_adam_2017}. Since expert and policy rollouts may diverge, we snapshot the simulator state before rollouts and restore it when generating expert labels to ensure consistent supervision.

We primarily train the RSC model with \(N=25\) UAVs, and train five models with prediction horizons \(H=1,10,20,30,40\).The formations for \(N=25\) is solid squares with five rows by five columns. The leader--follower target distance $R_t=5\,\mathrm{m}$.

The training data consist of 500 randomly generated maps with 3--6 obstacles. Obstacles are generated with a uniform distribution within a rectangular region centered at \((25,0)\) with length \(20\,\mathrm{m}\) and width \(30\,\mathrm{m}\), and the obstacle radius is \(1\,\mathrm{m}\text{--}2\,\mathrm{m}\).The initial positions of the UAVs are generated by applying a random perturbation, uniformly distributed within a \(5\,\mathrm{m}\) radius, to each agent's nominal position in the original formation.
The virtual target starts from the origin and moves along the positive \(x\)-axis at \(3\,\mathrm{m/s}\).

In the simulator, the communication radius \(R_c\) is set to \(6\,\mathrm{m}\), which is 1.2 times the leader--follower target distance. The maximum allowed acceleration is \(10\,\mathrm{m/s^2}\), and the maximum speed is \(10\,\mathrm{m/s}\). The cooldown time of Leader--Follower Reconfiguration is \(1\,\mathrm{s}\). The safety distance is \(0.5\,\mathrm{m}\); if the distance between a UAV and other UAVs or obstacle surfaces is smaller than the safety distance, the simulation is terminated early and marked as a failure. All simulations allow all UAVs up to 2100 steps (21s) to leave the obstacle region and complete reconfiguration with a maximum error smaller than \(10\%\); otherwise, it is marked as a failure.

We implement our model using PyTorch and the OpenAI gym framework in Python3.9. The server we use has Intel(R) Xeon(R) Platinum 8462Y+ CPU @ 4100MHz, NVIDIA A30 GPU and 24GB RAM.

To the best of our knowledge, we cannot find any paper that is highly related to our integrated setting. In the decentralized flocking field, we use two methods with similar large-scale flocking settings but without rigid formation settings as baselines. These two methods are the APF-based Olfati-Saber’s decentralized flocking algorithm and the imitation-learning-based STGNN decentralized flocking algorithm. For fairness, we add a formation potential field consistent with the expert to Olfati-Saber’s algorithm, add the observations required for rigid formation to the STGNN model, and conduct imitation learning with the same expert.

\subsection{Metrics}
\begin{enumerate}
    \item \textbf{Success Rate (\(S\%\)).} The probability that the UAV swarm successfully leaves the obstacle region and reconfigures into a rigid formation. The detailed definition of success is the same as that in Sec.~III-A; higher is better. The maximum formation error is defined as
    \begin{equation}
        \mathrm{ME}
        = \max_{i}\frac{\left|\left\lVert \mathbf{P}_{i\rightarrow \ell(i)}\right\rVert_{2}-R_t\right|}{R_t}.
    \end{equation}

    \item \textbf{Formation Time (\(\tau\)).} The time required for the UAV swarm to achieve \emph{Success}; shorter is better.

    \item \textbf{Mean Absolute Error (\(\mathrm{MAE}\)).} This metric measures the error between the predicted trajectories (before APF fusion) and the expert demonstrations. Although it lacks a direct correlation with the final task success rate, we retain it as an essential reference metric for imitation learning. To ensure a fair comparison across different prediction horizon lengths \(H\), we report the normalized metric \(\mathrm{MAE} = \mathrm{MAE}_{\mathrm{sum}} / H\). A smaller value indicates better prediction accuracy.

    \item \textbf{Non-collision Failure Rate} (\(D\%\)) All failures other than collisions are counted as non-collision failures. In our setting, non-collision failures mainly occur when UAVs get stuck around obstacles or when agents mutually block the occupation of target slots in the formation. Once such situations arise, they are likely to persist for the remainder of the episode, and thus can be understood as \emph{deadlock-like} behavior in decentralized coordination. Smaller is better.
\end{enumerate}

\subsection{Experimental Results}
We train one RSC model for each setting described in Section~\ref{sec:RSC Model Settings}. Each model is trained for 4000 epochs with a learning rate of \(3\times10^{-4}\).

For all tests, we use a newly generated test set consisting of 200 maps, where the map generation procedure is the same as that of the training set. The UAV positions are sampled using the same procedure as in the training set, but with different random seeds. For each model, we repeat the evaluation on the test set with three different seeds, resulting in 600 test runs in total, and report the mean value as the final result.

\paragraph{Evaluation on Model Performance}
In this set of experiment, five RSC models with different prediction horizon lengths \(H\) are evaluated. The update rule of these models is as follows: in the simulator, the system is updated entirely based on the model-predicted trajectories and the APF fusion results; after a total of \(H\) updates, the model produces another output. The Online Leader--Follower Reconfiguration mechanism is also enabled. In addition, the Expert is evaluated as the best reference, and two improved baselines are included for comparison. All methods are tested with 25 UAVs in a solid-square formation, and the results are reported in Table I.

\begin{table}[t]
\centering
\caption{TEST RESULTS OF $N = 25$. H DENOTES THE TRAJECTORY LENGTH. RSC-H20 ACHIEVES THE BEST PERFORMANCE.}
\label{tab:exp_n25}

\setlength{\tabcolsep}{7pt}
\renewcommand{\arraystretch}{1.15}
\small

\begin{tabular}{l|cccc}
\hline
\multicolumn{1}{c|}{Model} & \(S(\%)\) & \(\tau\)  & \(\mathrm{MAE}(1\times10^{-4})\) & \(D(\%)\) \\
\hline
Expert & 96.00  & 13.06 & -- & 3.00 \\
\hline
Saber  & 5.00 & 15.47 & -- & 39.50 \\
STGNN L3  & 0.00 & -- & -- & 72.00 \\
\hline
RSC-H1  & 0.00 & -- & 4.00 & 95.50 \\
RSC-H10 & 58.50 & \textbf{13.11} & 6.20 & 23.00 \\
RSC-H20 & \textbf{83.00} & 13.17 & 7.90 & \textbf{2.50} \\
RSC-H30 & 81.50 & 13.32 & 11.17 & 3.00 \\
RSC-H40 & 82.50 & 13.62 & 13.48 & \textbf{2.50} \\
\hline
\end{tabular}
\end{table}

In Table~\ref{tab:exp_n25}, we evaluate the swarm reconfiguration task with \(N=25\). Among all non-Expert methods, RSC-H20 achieves the best overall performance, attaining the highest success rate and the lowest non-collision failure rate. Meanwhile, its formation time is also very close to that of the best RSC-H10 model. From the perspective of success rate, the single-step controller Saber, STGNN L3, and the single-step trajectory predictor RSC-H1 all perform poorly. A plausible reason is that single-step methods tend to favor myopic, locally optimal decisions based on current observations, lacking mid- to long-horizon coordination for obstacle bypassing, yielding behavior, and target-slot assignment during formation.

As the prediction horizon \(H\) increases, the success rate improves significantly, and the RSC models with \(H=20,30,40\) exhibit comparable success-rate performance. Together with the non-collision failure rates, these results suggest that multi-step prediction is necessary in rigid swarm-formation scenarios. We argue that multi-step prediction partially captures the long-horizon planning of the centralized expert, mitigating the shortsightedness of single-step predictors and reducing non-collision failures to some extent.

Regarding MAE, it increases steadily with a longer prediction horizon, indicating reduced similarity to the expert under long-horizon predictions. We attribute this to two factors: (i) longer-horizon predictions lack intermediate observation-based corrections, making errors more likely to accumulate over time; and (ii) longer horizons require covering a richer set of behavior modes and scenario combinations, which increases learning difficulty and can degrade prediction accuracy.

\begin{figure}[t]
  \centering
  \includegraphics[width=\columnwidth]{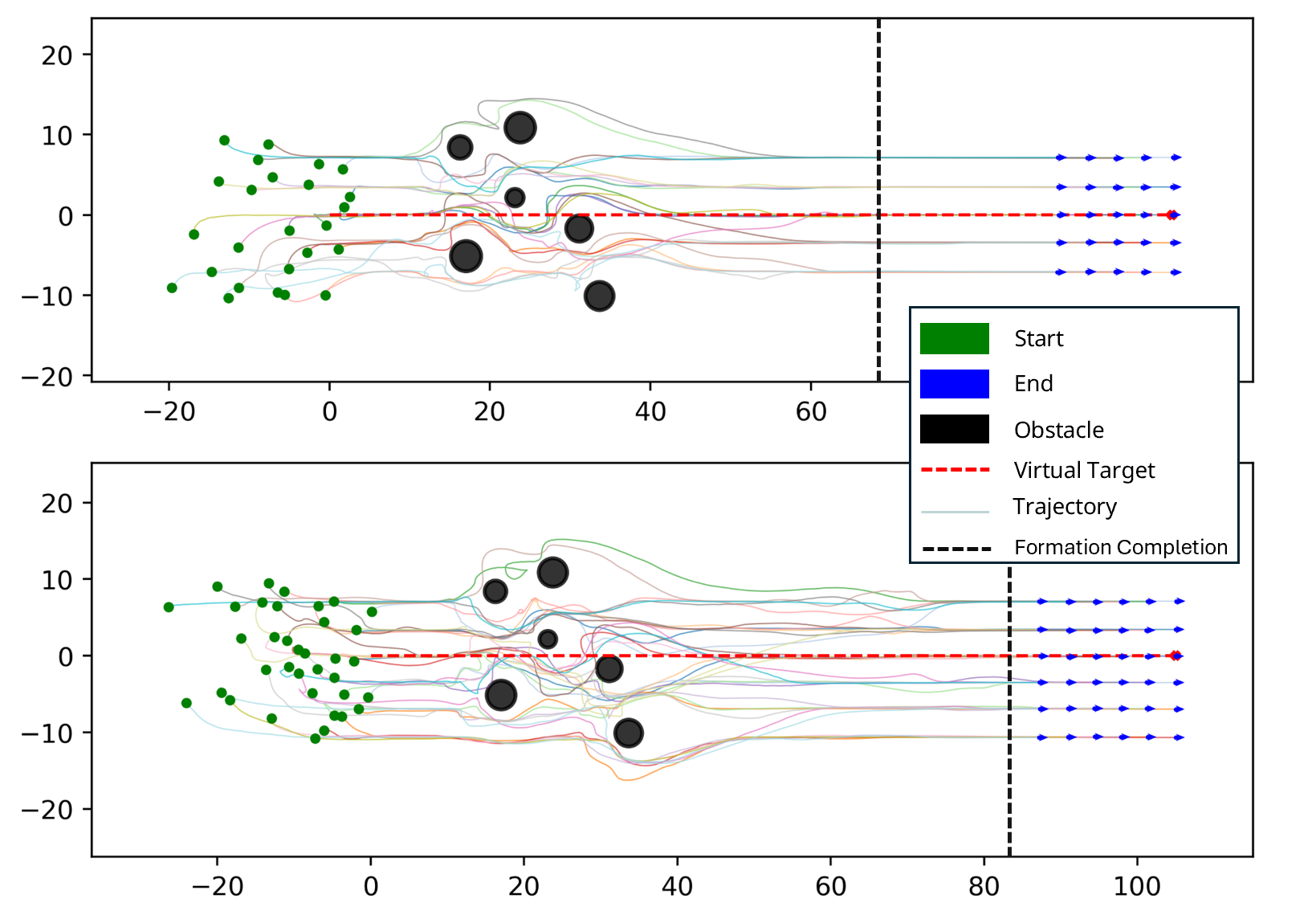}
  \caption{Simulation trajectories of the RSC-H20 model. The top and bottom panels demonstrate the navigation of 25 and 36 UAVs, respectively, within identical cluttered environments (the 9th map in the test set). The black dashed line marks the x-axis position of the rearmost UAV at the exact moment the formation is successfully reassembled.}
  \label{fig:small_teaser}
  \vspace{-10pt}
\end{figure}

\paragraph{Evaluation on Scalability and Transferability}
In these experiments, we evaluate scalability and transferability by training three separate models of the best-performing variants—RSC-H20, RSC-H30, and RSC-40—using a swarm size of \(N=25\). We then test each model on three formation configurations. For \(N=6\) and \(N=12\), the target formation is a hollow equilateral triangle; for \(N=36\), the target formation is a dense \(6\times6\) square grid. All other experimental settings are kept identical to those in the previous experiment. Notably, these swarm sizes (\(N>10\)) are unusually large in decentralized rigid formation control; to the best of our knowledge, prior work rarely reports evaluations at comparable scales. The results are summarized in Table~\ref{tab:exp_scala}.

\begin{table}[t]
\centering
\caption{Results under different swarm sizes \(N\).}
\label{tab:exp_scala}
\small
\setlength{\tabcolsep}{6pt}
\renewcommand{\arraystretch}{1.1}
\begin{tabular}{l|cccc}
\hline
\multicolumn{1}{c|}{} & \multicolumn{4}{c}{\textbf{\(N=6\)}} \\
Model & \(S(\%)\) & \(\tau\)  & \(\mathrm{MAE}(1\times10^{-4})\) & \(D(\%)\) \\
\hline
Expert & 97.00 & 9.51 & --     & 0.40 \\
RSC-H20     & 90.50 & 9.72 & 11.25 & 1.00 \\
RSC-H30     & \textbf{95.50} & \textbf{9.61} & 15.03 & \textbf{0.00} \\
RSC-H40     & 92.50 & 9.77 & 17.95 & 1.00 \\
\hline
\multicolumn{1}{c|}{} & \multicolumn{4}{c}{\textbf{\(N=12\)}} \\
Model & \(S(\%)\) & \(\tau\) & \(\mathrm{MAE}(1\times10^{-4})\) & \(D(\%)\) \\
\hline
Expert & 95.40 & 11.63 & --     & 0.00 \\
RSC-H20     & 90.00 & 11.70 & 7.70 & \textbf{0.00} \\
RSC-H30     & \textbf{90.50} & \textbf{11.55} & 13.83 & \textbf{0.00} \\
RSC-H40     & 87.50 & 11.74 & 11.01   & 2.00 \\
\hline
\multicolumn{1}{c|}{} & \multicolumn{4}{c}{\textbf{\(N=36\)}} \\
Model & \(S(\%)\) & \(\tau\) & \(\mathrm{MAE}(1\times10^{-4})\) & \(D(\%)\) \\
\hline
Expert & 89.80 & 15.17 & --     & 6.80 \\
RSC-H20     & 76.00 & \textbf{15.43} & 9.10 & \textbf{2.00} \\
RSC-H30     & \textbf{76.50} & 15.53 &  12.03   & 2.50 \\
RSC-H40     & 73.00 & 15.52 &   15.10   & 2.50 \\
\hline
\end{tabular}
\end{table}

As shown in Table~\ref{tab:exp_scala}, the results across different swarm sizes \(N\) exhibit a clear \emph{scaling effect}: as \(N\) increases from 6 to 36, the success rate \(S\) generally decreases while the completion time \(\tau\) increases, indicating that larger-scale decentralized formation control poses greater coordination and stability challenges. Nevertheless, although all learning-based models are trained only in the \(N=25\) setting, they remain effective when evaluated on both smaller swarms (\(N=6,12\)) and a larger swarm (\(N=36\)), demonstrating good \emph{transferability/generalization} to varying swarm sizes. Notably, in the large-scale case (\(N=36\)), the learned models are more advantageous in terms of avoiding \emph{deadlocks}: their deadlock rates \(D\) stay within \(2.0\%\sim2.5\%\), which is substantially lower than the Expert's \(6.8\%\). Overall, the three variants (RSC-20, RSC-30, and RSC-40) show no pronounced differences across swarm sizes; however, \emph{RSC-40 is slightly weaker}, as reflected by its lower success rates and the higher deadlock rate at \(N=12\), suggesting that this variant does not provide consistent benefits when scaling.

\subsection{Ablation Study}
In this set of experiments, we conduct ablation studies on two key components: APF fusion \((\mathcal{M}_1)\) and Online Leader--Follower Reconfiguration \((\mathcal{M}_2)\).
All experiments use the RSC-H20 model and follow the same settings as Experiment~a.
The results are reported in Table~\ref{tab:exp_abla}.

From the ablation study, we draw the following conclusions. First, \(\mathcal{M}_2\) (Online Leader--Follower Reconfiguration) is a key factor affecting performance. Removing \(\mathcal{M}_2\) causes the success rate to drop sharply from \(83.0\%\) to \(49.5\%\), while the non-collision failure rate \(D\) increases from \(2.5\%\) to \(34.0\%\), indicating that online L--F reconfiguration can effectively mitigate non-collision failures (e.g., deadlocks or stagnation caused by mutual yielding). Second, \(\mathcal{M}_1\) (APF fusion) provides a consistent gain: when \(\mathcal{M}_2\) is kept, removing \(\mathcal{M}_1\) reduces the success rate by \(4.5\) percentage points while \(D\) remains unchanged, suggesting that \(\mathcal{M}_1\) mainly helps reduce collision-related failures. Finally, removing both \(\mathcal{M}_1\) and \(\mathcal{M}_2\) yields the worst performance (\(S=38.0\%\), \(D=36.0\%\)), verifying that the two components are complementary. When the success rate is low (e.g., \(S=49.5\%\) or \(38.0\%\)), the values of \(\tau\) and \(\mathrm{MAE}\) are not strictly comparable: such models typically succeed only on easier maps, and failed episodes are not included in the mean statistics.
\begin{table}[t]
\centering
\caption{Ablation study}
\label{tab:exp_abla}
\begin{tabular}{cc|cccc}
\hline
\(\mathcal{M}_1\) & \(\mathcal{M}_2\) & \(S(\%)\) & \(\tau\) & \(\mathrm{MAE}(1\times10^{-4})\) & \(D(\%)\) \\
\hline
$\checkmark$  & $\checkmark$  & \textbf{83.00} & 13.17 & 7.90  & \textbf{2.50} \\
$\times$ & $\checkmark$  & 78.50    & 13.05    & 8.35    & \textbf{2.50}   \\
$\checkmark$  & $\times$ & 49.50    & 13.01    & 7.75    & 34.00   \\
$\times$ & $\times$ & 38.00    & \textbf{12.08}    & 8.75    & 36.00   \\
\hline
\end{tabular}
\end{table}

\subsection{Real robot experiment}
\begin{figure}[t]
  \centering
  \includegraphics[width=\columnwidth]{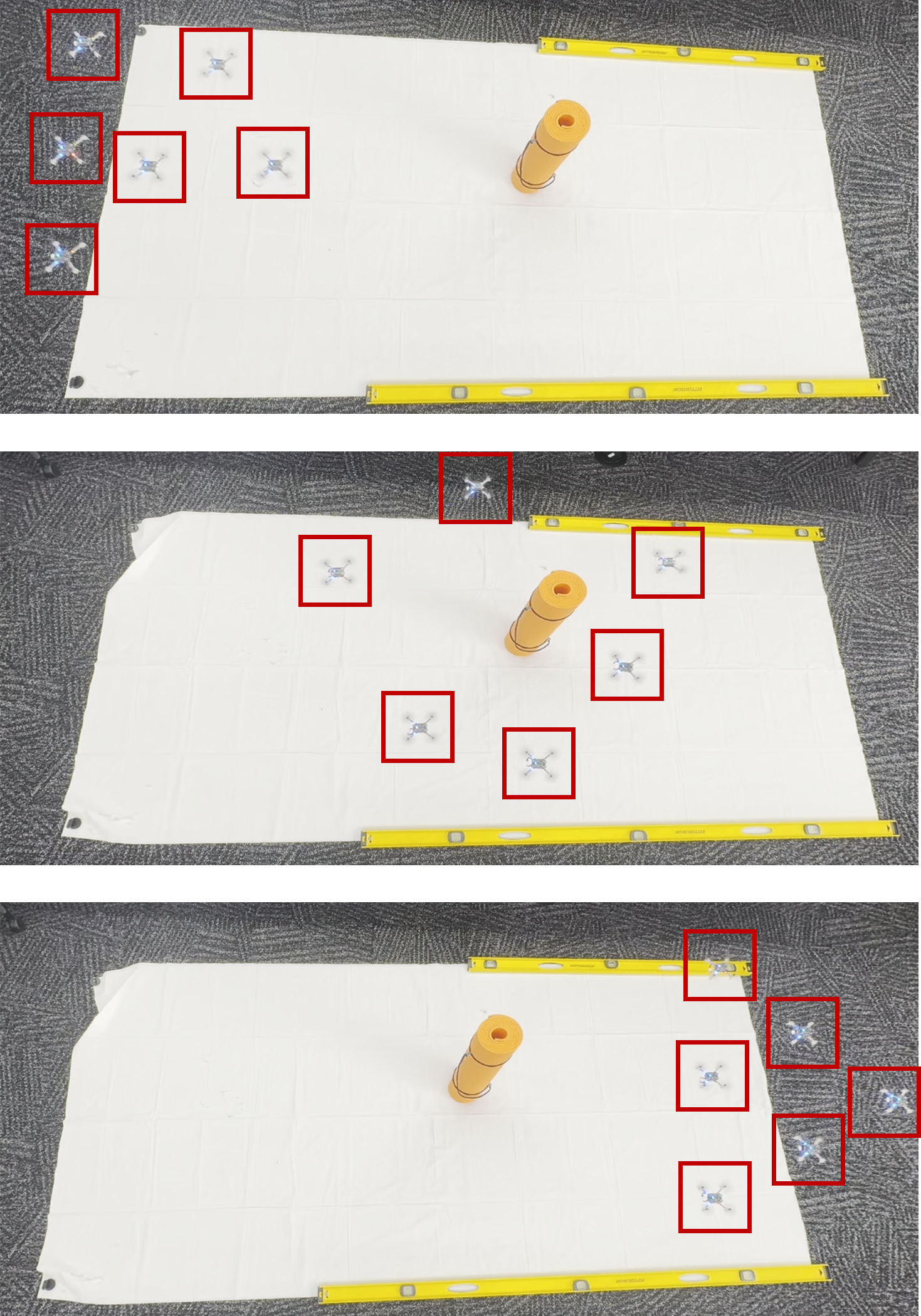}
  \caption{ The plots from top to bottom show an experiment of six drones
starting from random locations, avoiding an obstacle, and achieving a triangular formation.}
  \label{fig:4}
  \vspace{-10pt}
\end{figure}
We further demonstrate the effectiveness of RSC by applying it to achieve flocking behavior in a group of Bitcraze Crazyflie 2.1 drones. The drones are controlled via the Crazyswarm platform\cite{preiss_crazyswarm_2017}, which is built upon the Robot Operating System (ROS) and enables the vehicles to fly in tight, synchronized swarm formations. The positions of the drones are acquired via the LightHouse positioning system, which utilizes SteamVR base stations in conjunction with onboard positioning decks to estimate spatial coordinates. As illustrated in Figure~\ref{fig:4}, the six drones begin at random locations (top), navigate through the environment while avoiding obstacles (middle), and establish a triangular formation on the opposite side of the obstacles (bottom).
\section{CONCLUSIONS}
In this work, we proposed Rigid Swarm Control (RSC), a decentralized hybrid framework that successfully navigates the trade-off between long-term formation coordination and short-term collision avoidance. By synergizing finite-horizon predictive planning, reactive safety mechanisms, and dynamic topology reconfiguration, RSC effectively overcomes the deadlocks and local minima that typically plague large-scale swarm navigation in cluttered spaces.

Beyond the previously discussed missions, reliably maintaining and reassembling rigid spatial configurations at scale unlocks potential for physically interactive tasks. For instance, RSC can be applied to cooperative payload transport, where strict inter-agent distances are mandatory for load balance, or deployed as dynamic communication relay networks in confined subterranean environments. 

Future research will extend the RSC framework to highly dynamic environments with moving obstacles. Additionally, we plan to investigate the system's robustness under severely degraded communication channels (e.g., high packet loss and latency) and explore alternative formation paradigms that offer greater flexibility and efficiency than the traditional leader-follower topology.




\bibliographystyle{ieeetr}  
\bibliography{references}      

\end{document}